\documentclass[a4paper,french]{rnti}

\usepackage[T1]{fontenc}
\usepackage[utf8]{inputenc}

\usepackage{url}

\usepackage{graphicx}
\usepackage{color}

\titrecourt{Détection d'objets célestes dans des images astronomiques par IA explicable}

\titre{Détection d'objets célestes dans des images astronomiques par IA explicable}

\nomcourt{O. Parisot et al.}
\auteur{Olivier Parisot, Mahmoud Jaziri}
\affiliation{Luxembourg Institute of Science and Technology (LIST) \\ 5 Avenue des Hauts-Fourneaux L-4422 Luxembourg \\ olivier.parisot@list.lu}

\resume{%
Grâce à l'apport des télescopes automatisés grand public, les astronomes amateurs et professionnels peuvent capturer facilement une grande quantité d'images du ciel profond (comme par exemple les galaxies, nébuleuses, ou amas globulaires). Néanmoins, une vérification reste nécessaire à postériori pour vérifier si les objets célestes visés sont effectivement visibles dans les images produites: cela dépend notamment de la magnitude des cibles, des conditions d'observation mais aussi de la durée pendant laquelle les données sont capturées. Dans cette étude, nous proposons une approche basée sur l'IA explicable pour détecter automatiquement la présence et la position des objets capturés.
}

\summary{%
Amateur and professional astronomers can easily capture a large number of deep sky images with recent smart telescopes. However, afterwards verification is still required to check whether the celestial objects targeted are actually visible in the images produced. Depending on the magnitude of the targets, the observation conditions and the time during which the data is captured, it is possible that only stars are present in the images. In this study, we propose an approach based on explainable Artificial Intelligence to automatically detect the presence and position of captured objects.
}

\begin{document}

%
\section{Introduction}

Les télescopes intelligents (\textit{smart telescopes}) sont des dispositifs automatisés et disponibles pour le grand public, dédiés au Visuel Assisté (EAA en anglais, pour \textit{Electronically Assisted Astronomy}), et permettant des séances d'observation du ciel nocturne en famille et entre amis \citep{parisot2022improving}. 
Ils combinent des composants optiques, des caméras spécialisées et des montures de suivi pour capturer des images d'objets du ciel profond comme les galaxies, les nébuleuses et les amas globulaires. 
A travers les écrans de tablettes ou des smartphones, ils permettent d'admirer des cibles célestes peu lumineuses qui sont invisibles par observation directe \citep{varela2023increasing}: les capteurs sont beaucoup plus sensibles que l'œil humain pour capturer les signaux peu lumineux provenant de l'espace profond, et un traitement embarqué léger combine les images unitaires pour produire des images empilées de bonne qualité, même depuis des environnements touchés par la pollution lumineuse \citep{parker2007making}.

Les télescopes automatisés présentent également un intérêt scientifique évident: des collaborations récentes entre professionnels et amateurs ont montré que des cibles inconnues peuvent être découvertes en utilisant du matériel accessible aux amateurs, en accumulant des données sur une très longue période \citep{drechsler2023discovery}. 
En outre, l'utilisation simultanée d'un réseau de télescopes intelligents peut contribuer à l'étude des astéroïdes et même des exoplanètes \citep{peluso2023unistellar}.

\begin{figure}[!h]
        \center{\includegraphics[width=0.95\textwidth]{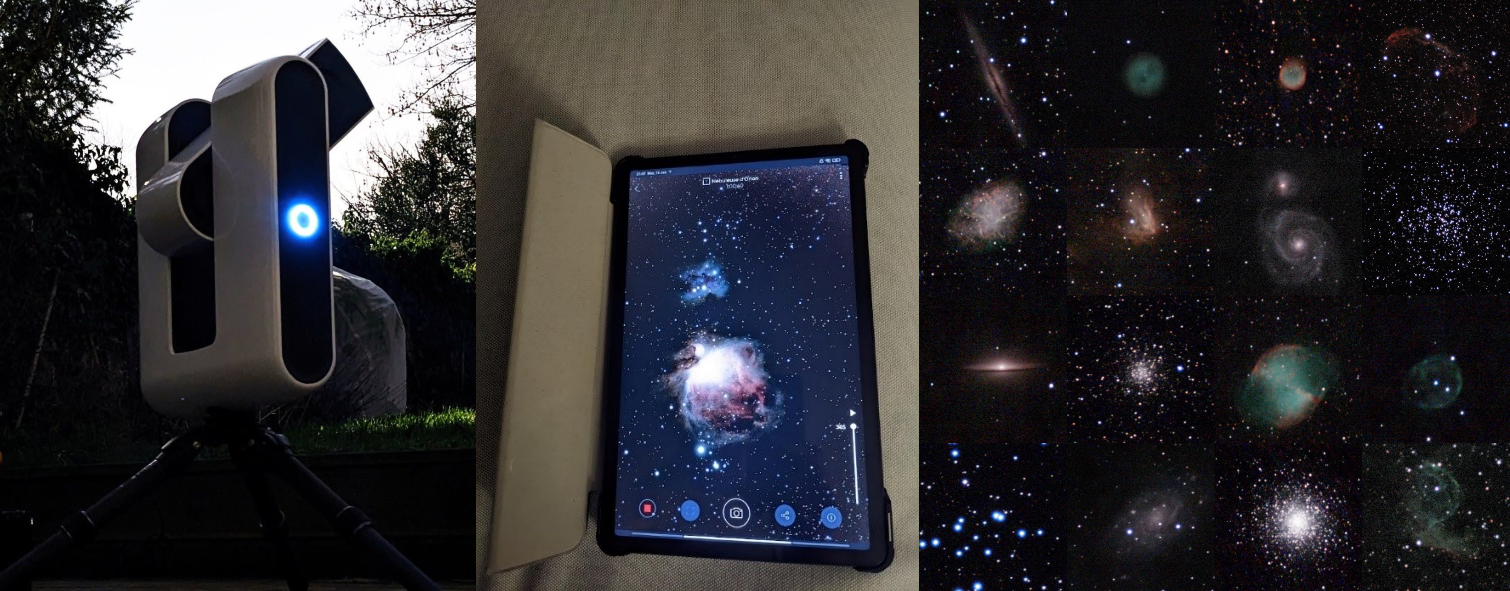}}
        \caption{Observation sur une tablette (au centre) de la Nébuleuse d'Orion avec un télescope Stellina (à gauche). A droite, un ensemble d'images astronomiques obtenues par les auteurs avec cet instrument.}
        \label{fig:stellina}
\end{figure}

En pratique, les télescopes intelligents permettent de programmer avec précision le démarrage et l'arrêt de la capture d'images d'une partie spécifique du ciel nocturne, et les résultats sont stockés pour une utilisation ultérieure sur un disque dur portable. 
C'est là qu'une fonction très importante peut s'avérer très utile: la détection d'objets qui sont effectivement visibles sur les images capturées. 
Si la présence d'étoiles sur les images ne fait guère de doute, il est plus difficile d'être certain d'avoir capturé une galaxie ou une nébuleuse, en particulier lorsque l'on vise des cibles difficiles de grande magnitude qui nécessitent de nombreuses heures (voire nuits)de capture. 
De plus, des conditions extérieures défavorables (pollution lumineuse, pleine lune, etc.) peuvent rendre difficile l'obtention d'images de qualité suffisante.
Il est également possible de capturer des objets qui n'étaient pas répertoriés \citep{drechsler2023discovery} ou attendus (exemple: comète, supernova). 
Il est donc utile de disposer d'un moyen permettant d'analyser automatiquement les images et de produire une image annotée avec les objets détectés.

Le reste de ce document est organisé comme suit. 
Premièrement, nous allons parler des techniques existantes pour détecter les objets dans les images astronomiques (Section \ref{sec:related}). 
Deuxièmement, nous présentons la méthode suivie pour capturer un ensemble d'images avec des télescopes automatisés (Section \ref{sec:data}). 
Troisièmement, nous détaillons deux approches pour détecter les objets dans ces images (Section \ref{sec:naive} et Section \ref{sec:approach}). 
Finalement nous concluons en discutant les résultats (Section \ref{sec:results}) puis en proposant des perspectives (Section \ref{sec:conclusion}).

\section{Etat de l'art}
\label{sec:related}

Traditionnellement, la détection d'objets astronomiques est réalisée en utilisant l'astrométrie (c'est-à-dire en trouvant la position exacte, l'échelle et l'orientation de l'image): en comparant avec les cartes du célestes connues (contenant les positions exactes des DSO), il est alors possible de trouver quels objets sont visibles sur l'image analysée \citep{lang2010astrometry}. 
En fait, l'astrométrie simplifiée / la résolution de plaques est utilisée lors de l'initialisation automatisée des télescopes intelligents -- de manière repérer les étoiles et donc l'orientation des instruments. 
C'est efficace, mais cela nécessite l'accès (local ou réseau) à une base de données contenant les coordonnées des corps célestes.
Et ces méthodes ne permettent pas de découvrir des objets encore non répertoriés.

Les approches de vision par ordinateur pour la détection d'objets sont également nombreuses, car elles permettent d'extraire des informations directement à partir des images. 
Il y a bientôt dix ans, un travail intéressant basé sur la segmentation a été proposé pour détecter des galaxies dans des relevés astronomiques \citep{zheng2015improved}. 
Récemment, plusieurs techniques basées sur l'intelligence artificielle (IA) ont été proposées. 
Parmi elles, les approches récentes sur YOLO (\emph{You Only Look Once}) sont spécialement dédiées à la détection d'objets dans les images, sur la base d'un entraînement supervisé préalable.
Par exemple, \citep{GONZALEZ2018103} propose de combiner YOLO et l'augmentation des données pour détecter et classer les types de galaxies dans les grands relevés astronomiques.
\citep{megacosm1_dataset} est un ensemble d'images représentant des corps célestes -- mais il contient trop peu d'images pour être utilisable tel quel (400).
Récemment, un papier a décrit comment détecter des coprs spatiaux à partir d'un jeu de donnée partiellement annoté \citep{dumitrescu2022novel}.

Ces méthodes basées sur l'IA nécessitent d'énormes ensembles de données d'entraînement pour être efficaces et, à notre connaissance, il n'existe pas de tel ensemble basé sur des images capturées avec du matériel accessible aux amateurs.
Dans ce papier, nous proposons une solution visant à traiter des images capturées avec des télescopes automatisés grand public, et qui nécessite un minimum d'étiquetage pour pouvoir détecter la présence et la position des objets.

\section{Données}
\label{sec:data}

Pour nos travaux, nous avons récolté une quantité conséquente de données (plus de 250 cibles différentes visibles depuis l'hémisphère nord). 
Les images ont été prises entre mars 2022 et septembre 2023 au Luxembourg, en France et en Belgique, en utilisant les fonctions d'alignement et d'empilement intégrées à ces deux instruments:
\begin{itemize}
\item Stellina \footnote{\url{https://vaonis.com/stellina}}: doublet ED avec une ouverture de 80 mm et une longueur focale de 400 mm (rapport focal de f/5) -- équipé d'un capteur CMOS Sony IMX178 d'une résolution de ~6,4 millions de pixels (3096 × 2080 pixels).
\item Vespera \footnote{\url{https://vaonis.com/vespera}}: quadruplet apochromatique avec une ouverture de 50 mm et une longueur focale de 200 mm (rapport focal de f/4) -- équipé d'un capteur CMOS Sony IMX462 d'une résolution de ~2 millions de pixels (1920 × 1080 pixels).
\end{itemize}

Pour chaque session d'observation, les instruments ont été installés dans un environnement sombre (sans lumière directe) et correctement équilibrés à l'aide d'un niveau à bulle sur un sol stable.
En fonction des conditions d'observation, des filtres CLS (\textit{City Light Suppression}) ou DB (\textit{Dual Band}) ont été utilisés pour obtenir davantage de signal sur les cibles à grande magnitude (en particulier pour les nébuleuses).
Les paramètres par défaut de Stellina et Vespera ont été appliqués: 10 secondes pour le temps d'exposition pour chaque  image unitaire et 20 dB pour le gain. 
Cette configuration est optimale: ce temps d'exposition est un bon compromis pour obtenir de bonnes images avec les montures motorisées alt-azimutales des instruments (une valeur plus élevée peut donner des traînées d'étoiles indésirables).
Les images ont été obtenues avec des temps d'intégration cumulés raisonnables (de 20 à 120 minutes - généralement suffisants pour obtenir un bon rapport signal-bruit pour la plupart des cibles): en astronomie, le temps d'intégration cumulé fait référence au temps total pendant lequel les données d'observation ont été capturées et traitées efficacement pour obtenir une image finale d'un objet ou d'une région du ciel.
Les données brutes sont disponibles depuis une archive ouverte \citep{parisot2023milan}.

\section{Approche naïve}
\label{sec:naive}

Une approche naïve consiste à ignorer les étoiles dans les images, pour ne considérer que ce que nous souhaitons détecter (c'est à dire les nébuleuses, galaxies, amas globulaires, etc.). 
Or, il n'est pas si facile de s'attaquer à cette tâche de manière systématique en utilisant les techniques conventionnelles de vision par ordinateur, car les étoiles n'ont pas toujours le même aspect (taille, couleur, halo), et elles peuvent se retrouver de manière apparente en avant-plan d'une galaxie ou d'une nébuleuse.

Récemment, des techniques d'IA ont été proposées pour traiter les images astronomiques \citep{kumar2022astronomy}, et la méthode la plus populaire pour les amateurs d'astrophotographie est StarNet \footnote{\url{https://www.starnetastro.com}}: elle consiste à supprimer les étoiles des images - produisant un effet visuel remarquable, en particulier sur les nébuleuses et les galaxies de grande taille apparente.
Ici, nous avons appliqué le modèle StarNet à notre ensemble d'images: nous avons ainsi obtenu deux versions de chaque image (avec et sans étoile), nous avons calculé les contours sur les images sans étoiles.

\begin{figure}[!h]
        \center{\includegraphics[width=\textwidth]{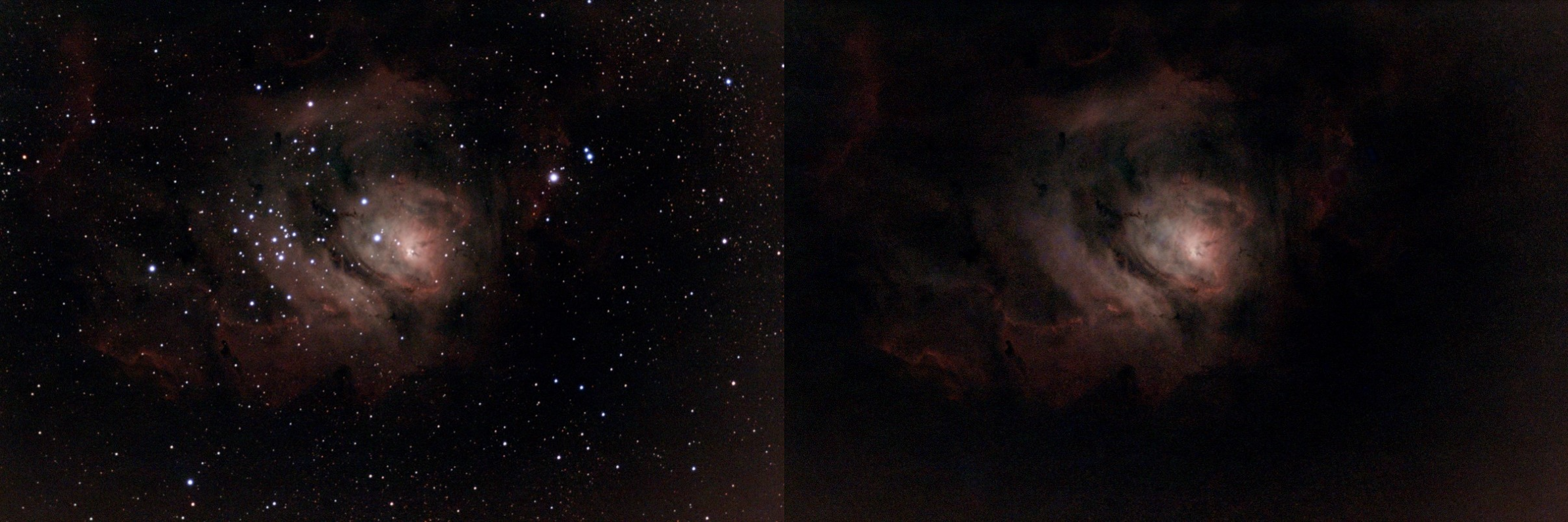}}
        \caption{Une version normale (à gauche) et une version \textit{starless} (à droite) d'une observation de la nébuleuse de la Lagune (Messier 8), capturée en août 2023 depuis un village de Haute Savoie avec un télescope Vespera. }
        \label{fig:starless}
\end{figure}

Cette méthode n'est pas parfaite pour discerner correctement tous les objets célestes: certaines galaxies sont supprimées par le modèle Starnet, et les faibles nébuleuses sont parfois confondues avec le fond de l'image (notamment lorsqu'il y a beaucoup de bruit).
C'est pourquoi nous essayons d'aller plus loin en proposant une approche originale pour détecter exclusivement ces objets.

\section{Apprentissage profond combinée avec IA explicable}
\label{sec:approach}

Inpiré par des récents travaux dans le domaine industriel \citep{roth2022towards} et dans le domaine de la santé \citep{chaddad2023survey}, notre approche consiste à entraîner un modèle de classification binaire pour détecter la présence des objets qui nous intéresse, puis d'appliquer une technique d'IA explicable pour identifier automatiquement leur position. 

L'IA explicable est un domaine de recherche actif qui vise à rendre interprétable les résultats d'un modèle IA.
De nos jours, ces outils sont considérés comme un outil de découverte scientifique \citep{li2022kepler}, en particulier pour comprendre les lois physiques en comparant les observations et les prédictions de l'IA \citep{roscher2020explainable}.

En pratique, voici les étapes suivies:
\begin{itemize}
\item Nous avons construit un ensemble de 5000 images RGB avec 224x224 pixels -- en appliquant des recadrages aléatoires pour obtenir des images de bonne taille.
\item Nous avons constitué deux groupes distincts, de manière à associer une classe à chaque image: les images avec et les images sans objets du ciel profond (nous avons veillé à ce que chaque groupe soit équilibré -- pour avoir un classifieur avec un bon rappel). Les images avec seulement des étoiles sont classées comme des images sans objets du ciel profond. Cette préparation a été réalisée en identifiant au préalable le type des objets ciblés dans les images grâce à Aladin \citep{bonnarel1999aladin}.
\item Nous avons fait 3 ensembles: entraînement, validation et test (80\%, 10\%, 10\%).
\item Un prototype Python dédié a été développé pour entraîner un modèle ResNet50 afin d'apprendre cette classification binaire. Les tâches de traitement d'image de base ont été réalisées en suivant les bonnes pratiques pour optimiser l'usage des CPU/GPU \citep{castro2023landscape}, et le prototype a été exécuté sur une infrastructure haute performance dotée des caractéristiques matérielles suivantes: 40 cœurs avec 128 Go de RAM (CPU Intel(R) Xeon(R) Silver 4210 @ 2,20 GHz) et NVIDIA Tesla V100-PCIE-32 Go.
\item De manière empirique, les hyperparamètres suivants ont été utilisés pendant l'entraînement: optimiseur ADAM, taux d'apprentissage de 0.001, 50 époques, 16 images par batch. Nous avons ainsi obtenu un modèle ResNet50 ayant une précision de 97\% sur l'ensemble de données de validation. Précisons que les architectures VGG16 et MobileNetV2 ont également été testées, mais les résultats ici sont en grande partie similaires.
\item Pour l'inférence des résultats, nous avons construit un pipeline pour analyser la sortie du modèle ResNet50 entraîné avec XRAI (\textit{Region-based Image Attribution}) \citep{kapishnikov2019xrai}. XRAI est une méthode incrémentale qui construit progressivement les régions d'attribution (i.e. les régions de l'image les plus importantes pour la classification) et qui fournit de bons résultats sur les images sombres. En pratique, nous avons utilisé le package Python \emph{saliency} \footnote{\url{https://pypi.org/project/saliency/}} et nous avons analysé la sortie de la dernière couche de convolution. Nous générons ainsi une \textit{heatmap} indiquant les régions d'attribution avec le plus grand pouvoir prédictif.
\end{itemize}


En combinant Resnet50 et XRAI, nous pouvons estimer la position des objets visibles dans des images astronomiques, sans avoir eu à annoter précisément la position de ces objets pour la phase d'entraînement du modèle.
Pour traiter une image haute résolution, nous découpons l'image en patch 224x224 \footnote{\url{https://pypi.org/project/patchify/}}, Nous appliquons le pipeline Resnet50 et XRAI sur chaque patch puis nous reconstituons le tout pour obtenir la même taille que l'image d'origine.

\section{Résultats et discussion}
\label{sec:results}


Si le modèle ResNet50 trouve qu'une image contient effectivement des objets, alors la \textit{heatmap} produite par XRAI localise les zones de l'image qui conduisent à ce résultat.

\begin{figure}[!h]
        \center{\includegraphics[width=\textwidth]{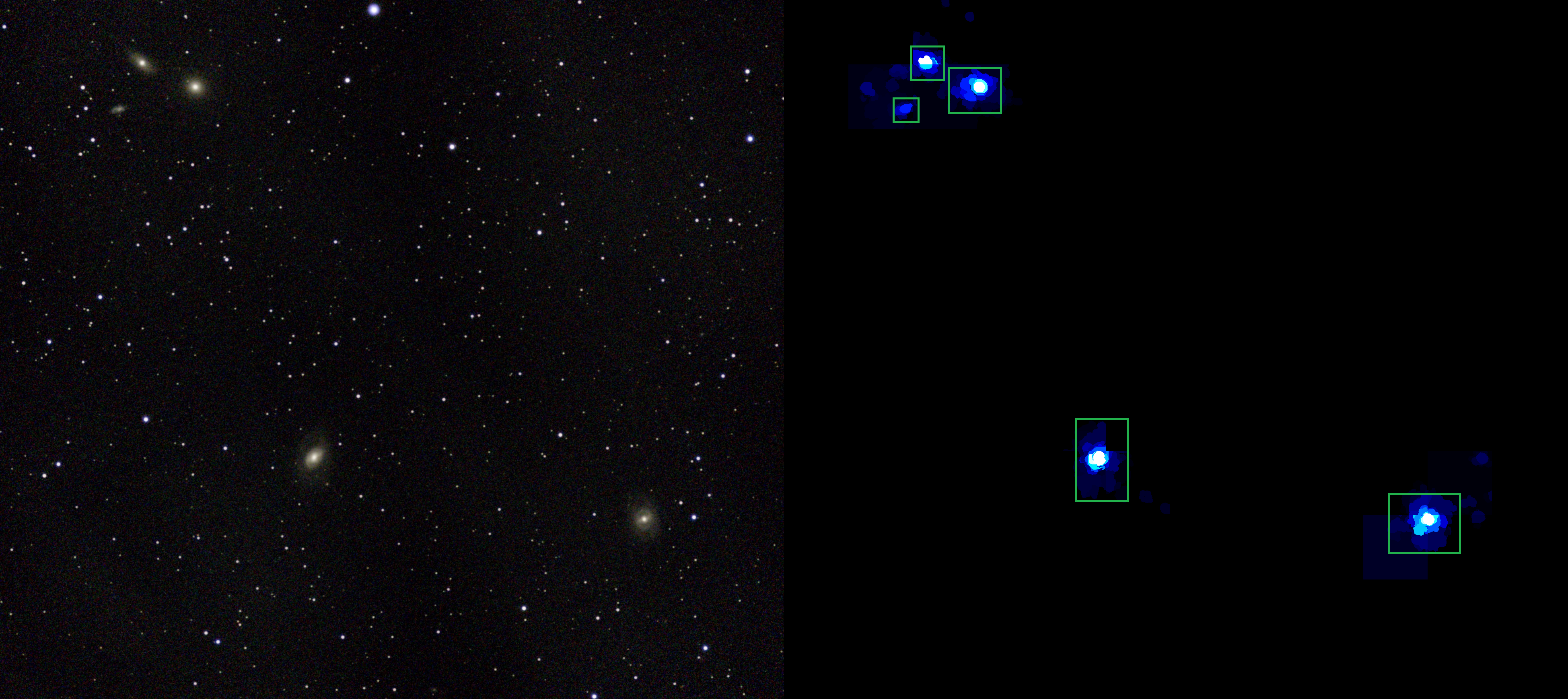}}
        \caption{A gauche, une image d'un champ de vision contenant plusieurs galaxies (dont Messier 95 et Messier 96) capturées avec un télescope intelligent Vespera. À droite, la heatmap résultant du pipeline ResNet50+XRAI, et les contours verts qui en découlent.}
        \label{fig:m95}
\end{figure}

Pour estimer concrètement l'efficacité de la méthode:
\begin{itemize}
\item Nous avons sélectionné un jeu de données représentatif de 100 images haute résolution (mixant nébuleuses, galaxies et amas globulaires) puis nous les avons annoté \textit{manuellement} (i.e. dessiné les contours) avec l'outil MakeSense \citep{makesense}.
\item Nous avons ensuite exécuté le modèle ResNet50 puis appliqué XRAI pour obtenir les \textit{heatmap} de chaque image de cet ensemble, pour ensuite calculer les contours correspondants en utilisant openCV.
\item Nous avons enfin comparé les contours dessinés avec MakeSense et les contours calculés à partir de la \textit{heatmap} XRAI, avec le package Python mAP \citep{8594067} (\footnote{\url{https://github.com/Cartucho/mAP}}), et nous avons ainsi calculé que la seconde approche a une précision de 0.79, un rappel de 0.41 et une mAP (\textit{mean average precision}) de 0.75.
\end{itemize}

Par comparaison, appliquer le modèle Starnet puis calculer les contours avec openCV est peu efficace car il reste toujours des résidus dans les images \textit{starless} qui font que les contours résultants sont moins satisfaisants (précision de 0.45, rappel de 0.36), en particulier pour les larges nébuleuses.


Un autre point concerne le temps de calcul sur des images à haute résolution.
Appliquer XRAI a un coût en temps de calcul et en ressources qui n'est pas négligeable, cela nécessite plus de ressources qu'une simple inférence du modèle ResNet50.
Prenons l'exemple d'une image astronomique de 3584x3584: sans chevauchement, il peut être nécessaire d'évaluer la prédiction ResNet50 et la \textit{heatmap} XRAI pour 256 patchs de 224x224 -- cela peut prendre un certain temps en fonction du matériel.
Pour être efficace, il faut essayer de minimiser le nombre de calculs nécessaires.
De manière pragmatique, ces stratégies peuvent être appliquées:
\begin{itemize}
\item Réduire la taille de l'image pour diminuer le nombre de patchs à évaluer.
\item Ne traiter qu'un sous-ensemble pertinent de patchs -- par exemple en ignorant ceux pour lesquels le classificateur ResNet50 ne détecte rien.
\end{itemize}

Au cours de nos expériences, nous avons observé que la seconde stratégie donnait de bons résultats.
D'autres optimisations des performances seront réalisées après une analyse approfondie de l'exécution du modèle à l'aide d'outils dédiés \citep{9150136}.

\section{Conclusion}
\label{sec:conclusion}

Cet article présente une approche permettant de détecter la présence et la position d'objects célestes dans des images astronomiques capturées avec du matériel disponible pour le grand public. 
Nous avons capturé une grande quantité d'images, nous les avons divisé en deux groupes distincts (avec et sans objets), puis nous avons entraîné un classifieur ResNet50 via un prototype développé en Python.
A l'aide de XRAI, nous avons mis au point un pipeline permettant de déterminer avec une précision acceptable les contours des objets visibles, notamment par rapport à une méthode basée sur un modèle d'extraction des étoiles puis de recherche de contours.
Cette technique permet de compléter les outils existants d'astrométrie pour permettre l'identification de corps célestes encore non répertoriés (comètes, supernova, etc.).

Dans nos futurs travaux, nous continuerons à capturer des images puis à les traiter pour produire un jeu de données YOLO, puis nous viserons d'appliquer d'autres techniques originales pour annoter les images automatiquement, notamment à base d'IA générative.



\bibliographystyle{rnti}
\bibliography{references}

\providecommand\Fr{}
\providecommand\Eng{}
\providecommand\andname{and}
\providecommand\andnamec{and}

\begin{thebibliography}{}


\bibitem[{Bonnarel et~al.}(1999){Bonnarel, Fernique, Genova, Bartlett,
  Bienaym{\'e}, Egret, Florsch, Ziaeepour, \andnamec{}
  Louys}]{bonnarel1999aladin}
Bonnarel, F., P.~Fernique, F.~Genova, J.~G. Bartlett, O.~Bienaym{\'e},
  D.~Egret, J.~Florsch, H.~Ziaeepour, \andname{} M.~Louys (1999).
\newblock Aladin: A reference tool for identification of astronomical sources.
\newblock In {\em Astronomical Data Analysis Software and Systems VIII}, Volume
  172, pp.\  229.

\bibitem[{{Cartucho} et~al.}(2018){{Cartucho}, {Ventura}, \andnamec{}
  {Veloso}}]{8594067}
{Cartucho}, J., R.~{Ventura}, \andname{} M.~{Veloso} (2018).
\newblock Robust object recognition through symbiotic deep learning in mobile
  robots.
\newblock In {\em 2018 IEEE/RSJ International Conference on Intelligent Robots
  and Systems (IROS)}, pp.\  2336--2341.

\bibitem[{Castro et~al.}(2023){Castro, Bruneau, Sottet, \andnamec{}
  Torregrossa}]{castro2023landscape}
Castro, O., P.~Bruneau, J.-S. Sottet, \andname{} D.~Torregrossa (2023).
\newblock Landscape of high-performance python to develop data science and
  machine learning applications.
\newblock ~{\em 56\/}(3).

\bibitem[{Chaddad et~al.}(2023){Chaddad, Peng, Xu, \andnamec{}
  Bouridane}]{chaddad2023survey}
Chaddad, A., J.~Peng, J.~Xu, \andname{} A.~Bouridane (2023).
\newblock {Survey of explainable AI techniques in healthcare}.
\newblock {\em Sensors\/}~{\em 23\/}(2), 634.

\bibitem[{Drechsler et~al.}(2023){Drechsler, Strottner, Sainty, Fesen,
  Kimeswenger, Shull, Falls, Vergnes, Martino, \andnamec{}
  Walker}]{drechsler2023discovery}
Drechsler, M., X.~Strottner, Y.~Sainty, R.~A. Fesen, S.~Kimeswenger, J.~M.
  Shull, B.~Falls, C.~Vergnes, N.~Martino, \andname{} S.~Walker (2023).
\newblock {Discovery of Extensive [O iii] Emission Near M31}.
\newblock {\em Research Notes of the AAS\/}~{\em 7\/}(1), 1.

\bibitem[{Dumitrescu et~al.}(2022){Dumitrescu, Ceachi, Truic{\u{a}},
  Tr{\u{a}}sc{\u{a}}u, \andnamec{} Florea}]{dumitrescu2022novel}
Dumitrescu, F., B.~Ceachi, C.-O. Truic{\u{a}}, M.~Tr{\u{a}}sc{\u{a}}u,
  \andname{} A.~M. Florea (2022).
\newblock A novel deep learning-based relabeling architecture for space objects
  detection from partially annotated astronomical images.
\newblock {\em Aerospace\/}~{\em 9\/}(9), 520.

\bibitem[{González et~al.}(2018){González, Muñoz, \andnamec{}
  Hernández}]{GONZALEZ2018103}
González, R., R.~Muñoz, \andname{} C.~Hernández (2018).
\newblock Galaxy detection and identification using deep learning and data
  augmentation.
\newblock {\em {Astronomy and Computing}\/}~{\em 25}, 103--109.

\bibitem[{Jin \andnamec{} Finkel}(2020){Jin \andnamec{} Finkel}]{9150136}
Jin, Z. \andname{} H.~Finkel (2020).
\newblock Analyzing deep learning model inferences for image classification
  using {OpenVINO}.
\newblock In {\em IPDPSW 2020}, pp.\  908--911.

\bibitem[{Kapishnikov et~al.}(2019){Kapishnikov, Bolukbasi, Vi{\'e}gas,
  \andnamec{} Terry}]{kapishnikov2019xrai}
Kapishnikov, A., T.~Bolukbasi, F.~Vi{\'e}gas, \andname{} M.~Terry (2019).
\newblock {XRAI: Better attributions through regions}.
\newblock In {\em Proceedings of the IEEE/CVF International Conference on
  Computer Vision}, pp.\  4948--4957.

\bibitem[{Kumar}(2022){Kumar}]{kumar2022astronomy}
Kumar, A. (2022).
\newblock {Astronomy and AI Beyond conventional astronomy}.
\newblock \url{ https://ir.iimcal.ac.in:8443/jspui/handle/123456789/4067 }.
\newblock visited on 2023-11-11.

\bibitem[{Lang et~al.}(2010){Lang, Hogg, Mierle, Blanton, \andnamec{}
  Roweis}]{lang2010astrometry}
Lang, D., D.~W. Hogg, K.~Mierle, M.~Blanton, \andname{} S.~Roweis (2010).
\newblock Astrometry. net: Blind astrometric calibration of arbitrary
  astronomical images.
\newblock {\em The astronomical journal\/}~{\em 139\/}(5), 1782.

\bibitem[{Li et~al.}(2022){Li, Ji, \andnamec{} Zhang}]{li2022kepler}
Li, Z., J.~Ji, \andname{} Y.~Zhang (2022).
\newblock {From Kepler to Newton: Explainable AI for Science Discovery}.
\newblock In {\em ICML 2022 2nd AI for Science Workshop}.

\bibitem[{Parisot et~al.}(2022){Parisot, Bruneau, Hitzelberger, Krebs,
  \andnamec{} Destruel}]{parisot2022improving}
Parisot, O., P.~Bruneau, P.~Hitzelberger, G.~Krebs, \andname{} C.~Destruel
  (2022).
\newblock Improving accessibility for deep sky observation.
\newblock {\em ERCIM News\/}~{\em 2022\/}(130).

\bibitem[{Parisot et~al.}(2023){Parisot, Hitzelberger, Bruneau, Krebs,
  Destruel, \andnamec{} Vandame}]{parisot2023milan}
Parisot, O., P.~Hitzelberger, P.~Bruneau, G.~Krebs, C.~Destruel, \andname{}
  B.~Vandame (2023).
\newblock {MILAN Sky Survey, a dataset of raw deep sky images captured during
  one year with a Stellina automated telescope}.
\newblock {\em Data in Brief\/}~{\em 48}, 109133.

\bibitem[{Parker}(2007){Parker}]{parker2007making}
Parker, G. (2007).
\newblock {\em {Making Beautiful Deep-Sky Images}}.
\newblock Springer.

\bibitem[{Peluso et~al.}(2023){Peluso, Esposito, Marchis, Dalba, Sgro,
  Megowan-Romanowicz, Pennypacker, Carter, Wright, Avsar,
  et~al.}]{peluso2023unistellar}
Peluso, D.~O., T.~M. Esposito, F.~Marchis, P.~A. Dalba, L.~Sgro,
  C.~Megowan-Romanowicz, C.~Pennypacker, B.~Carter, D.~Wright, A.~M. Avsar,
  et~al. (2023).
\newblock The unistellar exoplanet campaign: Citizen science results and
  inherent education opportunities.
\newblock {\em Publications of the Astronomical Society of the Pacific\/}~{\em
  135\/}(1043), 015001.

\bibitem[{Priyanka}(2022){Priyanka}]{megacosm1_dataset}
Priyanka (2022).
\newblock megacosm1 dataset.
\newblock \url{ https://universe.roboflow.com/priyanka-1uoyf/megacosm1 }.
\newblock visited on 2023-11-11.

\bibitem[{Roscher et~al.}(2020){Roscher, Bohn, Duarte, \andnamec{}
  Garcke}]{roscher2020explainable}
Roscher, R., B.~Bohn, M.~F. Duarte, \andname{} J.~Garcke (2020).
\newblock Explainable machine learning for scientific insights and discoveries.
\newblock {\em Ieee Access\/}~{\em 8}, 42200--42216.

\bibitem[{Roth et~al.}(2022){Roth, Pemula, Zepeda, Sch{\"o}lkopf, Brox,
  \andnamec{} Gehler}]{roth2022towards}
Roth, K., L.~Pemula, J.~Zepeda, B.~Sch{\"o}lkopf, T.~Brox, \andname{} P.~Gehler
  (2022).
\newblock Towards total recall in industrial anomaly detection.
\newblock In {\em Proceedings of the IEEE/CVF Conference on Computer Vision and
  Pattern Recognition}, pp.\  14318--14328.

\bibitem[{Skalski}(2019){Skalski}]{makesense}
Skalski, P. (2019).
\newblock {Make Sense}.
\newblock \url{https://github.com/SkalskiP/make-sense/}.

\bibitem[{Varela~Perez}(2023){Varela~Perez}]{varela2023increasing}
Varela~Perez, A.~M. (2023).
\newblock The increasing effects of light pollution on professional and amateur
  astronomy.
\newblock {\em Science\/}~{\em 380\/}(6650), 1136--1140.

\bibitem[{Zheng et~al.}(2015){Zheng, Pulido, Thorman, \andnamec{}
  Hamann}]{zheng2015improved}
Zheng, C., J.~Pulido, P.~Thorman, \andname{} B.~Hamann (2015).
\newblock An improved method for object detection in astronomical images.
\newblock {\em Monthly Notices of the Royal Astronomical Society\/}~{\em
  451\/}(4), 4445--4459.

\end{thebibliography}




\end{document}